\documentclass[conference]{IEEEtran}
\IEEEoverridecommandlockouts
\usepackage{cite}
\usepackage{amsmath,amssymb,amsfonts}
\usepackage{algorithmic}
\usepackage{graphicx}
\usepackage{textcomp}
\usepackage{xcolor}
\def\BibTeX{{\rm B\kern-.05em{\sc i\kern-.025em b}\kern-.08em
    T\kern-.1667em\lower.7ex\hbox{E}\kern-.125emX}}
\begin{document}

\title{Hybrid Voting-Based Task Assignment in Role-Playing Games\\
}

\author{\IEEEauthorblockN{{Daniel Weiner}}
\IEEEauthorblockA{\textit{Computer Science Department} \\
\textit{Graduate Center, City University of New York}\\
New York, United States \\
dweiner@gradcenter.cuny.edu}
\and
\IEEEauthorblockN{Raj Korpan}
\IEEEauthorblockA{\textit{Computer Science Department} \\
\textit{Hunter College and Graduate Center, City University of New York}\\
New York, United States \\
raj.korpan@hunter.cuny.edu}
}

\maketitle

\begin{abstract}

In role-playing games (RPGs), the level of immersion is critical–especially when an in-game agent conveys tasks, hints, or ideas to the player. For an agent to accurately interpret the player’s emotional state and contextual nuances, a foundational level of understanding is required, which can be achieved using a Large Language Model (LLM). Maintaining the LLM’s focus across multiple context changes, however, necessitates a more robust approach, such as integrating the LLM with a dedicated task allocation model to guide its performance throughout gameplay. In response to this need, we introduce Voting-Based Task Assignment (VBTA), a framework inspired by human reasoning in task allocation and completion. VBTA assigns capability profiles to agents and task descriptions to tasks, then generates a suitability matrix that quantifies the alignment between an agent’s abilities and a task’s requirements. Leveraging six distinct voting methods, a pre-trained LLM, and integrating conflict-based search (CBS) for path planning, VBTA efficiently identifies and assigns the most suitable agent to each task. While existing approaches focus on generating individual aspects of gameplay, such as single quests, or combat encounters, our method shows promise when generating both unique combat encounters and narratives because of its generalizable nature.
\end{abstract}


\section{Introduction}

Procedural Content Generation (PCG) in role-playing games (RPGs) has not addressed longer narratives and connected storylines, instead specializing in generating or assisting with individual aspects of gameplay \cite{Survey}. Immersion, the seamless integration of narrative, emotional engagement, and real-time responsiveness, is critical for maintaining user engagement. This requires crafting an all-encompassing experience that transports players into an alternative world where they feel as if they have become the character they are playing, and that they are a genuine part of this fantasy. Current generative PCG methods do not achieve this as they only manage one aspect of gameplay at a time \cite{Survey}. To address this challenge, we propose Voting-Based Task Assignment (VBTA), a novel framework inspired by human reasoning processes for task allocation and completion that couples a structured task allocation model with LLM-driven semantic interpretation. This paper proposes that by integrating structured task allocation with LLM-driven semantic interpretation, the VBTA framework can overcome the limitations of current generative PCG methods, enabling the creation of immersive, cohesive, and continuously evolving RPG experiences. We present use cases with Dungeons \& Dragons and Baldur's Gate 3 that demonstrate VBTA's potential. 

In recent years, the pursuit of immersive experiences in role-playing games and human-robot interaction (HRI) has driven significant advances in how autonomous agents engage with human users \cite{SceneCraft, 1001Nights, Quests1, Quests2, Calypso}. Immersion is highly dependent on the narrative and the agents in any RPG. Narrative in RPGs refers to the overarching story that shapes the game world, including its plot, lore, character arcs, and events that unfold over the course of game play, dialogue, on the other hand, is the direct verbal exchange between characters that make up part of the narrative framework. Agents are any intelligent entities that populate the game world, including NPCs, enemies, allies, and other creatures. They are designed to operate autonomously or semi-autonomously, making decisions and interacting with both players and the environment to enrich gameplay, drive the narrative, and create a dynamic experience. In gaming environments, agents must deliver tasks, hints, or narrative cues and interpret and adapt to the player’s emotional state and situational context. Context switching is the act of changing between different tasks in a game, such as two unrelated dialogues or quests. Context switching presents a challenge to PCG methods attempting to generate content for a game. Traditional approaches to agent communication often struggle to maintain consistency and relevance \cite{SceneCraft, 1001Nights, Quests1, Quests2}, particularly when faced with frequent context switches \cite{Calypso}. Recent work with LLMs has demonstrated their potential for deep narrative understanding and applying that insight to the characters in a scene \cite{SceneCraft}. However, while LLMs provide a robust foundation for interpreting nuanced human language, their performance in dynamic, multi-task, multi-agent scenarios can be inconsistent if not guided by a complementary decision-making framework \cite{SceneCraft}.

In the VBTA framework, each agent is associated with a capability profile, and each task is defined by a clear set of requirements known as a task description. By generating a suitability matrix that scores the alignment between agents and tasks, VBTA ensures that the most qualified agent is identified for each task. Recognizing that the highest-scoring agent may be best suited for multiple tasks, our approach incorporates six distinct voting methods alongside three allocation strategies to balance the best agent assignments across all available tasks. Furthermore, in cases where ambiguity arises–such as when an agent’s capability profile and a task description do not indicate clear compatibility–a pre-trained LLM is leveraged to resolve these uncertainties by inferring deeper semantic connections. Custom prompts are generated to query the LLM and ensure that the agent is capable of completing the task regardless of the uncertain compatibility, allowing more flexible allocation than classical voting. Once task assignments are finalized, conflict-based search (CBS) \cite{CBS} is employed for path planning, ensuring that agents not only receive appropriate tasks but can also navigate efficiently within complex game scenarios. This integration of semantic reasoning with formal task and path assignment algorithms represents a significant step toward more adaptable and human-like agent behaviors. We anticipate that the VBTA framework will serve as a hybrid approach by integrating voting with semantic reasoning from LLMs and dynamic game state planning via CBS for agent action decisions, generate affect-responsive game content spanning narrative, dialogue, and combat scenarios, and automate decisions that have previously required external intervention, thereby addressing several limitations of current methods.

\section{Related Work}

Recent advances in LLMs have spurred a variety of approaches to procedural content generation and narrative design in games. These methods can broadly be categorized into level generation, narrative scene creation, and RPG assistance, each with its own strengths and limitations.

\subsection{Level Generation With LLMs}
MarioGPT \cite{MarioGPT} is an early example that employs a GPT-2 \cite{GPT2} model to generate Super Mario Bros levels. By combining text-to-level generation with a novelty search algorithm, MarioGPT can produce diverse levels in an open-ended manner. Although 88.4\% of the generated levels were deemed playable by a powerful Super Mario algorithm, the system struggles to strictly adhere to design prompts. Its generated levels often feature inaccuracies in element counts (such as pipes, enemies, blocks, and elevation), and the model tends to memorize and regurgitate data rather than innovate unique experiences. Similar challenges are discussed in work on level generation through LLMs, where issues of spatial token representation and data scarcity are prominent \cite{Sokoban}. A related study \cite{Metavoidal} addresses level generation for a more complex and recent game (Metavoidal) by engineering constraint-based prompts for GPT-3 \cite{GPT3}. Despite these efforts, the approach achieved only a 37\% playable level generation rate and required significant human-in-the-loop augmentation, highlighting the difficulties of scaling LLM-driven level design to larger, more complex environments.

\subsection{Narrative Scene Generation}
SceneCraft \cite{SceneCraft} represents another line of work that targets interactive narrative scene generation. This framework transforms an author's written story summary into a choose-your-own-adventure–style interaction by using an LLM to generate narrative scenes. SceneCraft also extracts emotions and gestures from the generated text to enrich non-player character (NPC) interactions. However, while SceneCraft is innovative in extracting emotions and gestures, it is limited by its focus on creating short, isolated narrative scenes rather than an overarching narrative structure. Its focus on single, self-contained scenes means that it lacks a higher-level narrative planner to bind these scenes into a cohesive larger narrative. A study of the game 1001 Nights expressed a similar struggle with keeping the entirely generated responses of the King character on task \cite{1001Nights}. With simple prompting to the LLM, a player could force the King to break, ruining player immersion and the narrative of Persian folklore that the game was set in \cite{1001Nights}.   

\subsection{RPG Assistance and Dungeon Master Tools}
In the domain of role-playing games, Calypso \cite{Calypso} offers an LLM-based Dungeon Master (DM) assistant with interfaces tailored to support Dungeons \& Dragons (D\&D) gameplay. Its modules focus mainly on encounters, assisting with enemy information and battle narratives. Although Calypso excels at setting up and refining encounters, its general conversational interface is standard ChatGPT \cite{chatgpt} and does not extend to real-time narrative creation. While other approaches focus on generating RPG quests and dialogues using GPT models \cite{SceneCraft, 1001Nights}, they all similarly address individual components of gameplay in small digestible chunks that are easy for an LLM to comprehend. Without a framework to manage the context, the LLM has trouble generating expansive narratives.


\section{Voting-Based Task Assignment}

The VBTA framework adopts a hybrid approach that integrates structured task descriptions and capability profiles with semantic reasoning from LLMs and dynamic path planning from CBS to automate agent actions and generate comprehensive game content spanning narrative dialogue, combat scenarios, and environmental interactions. By automating decisions that previously required external intervention, VBTA addresses key limitations of earlier approaches and enables the creation of rich, continuously evolving game worlds with high replayability and deep player immersion. VBTA integrates detailed \textit{Task Descriptions} with comprehensive agent \textit{Capability Profiles} to facilitate effective agent and task assignments. Each task is defined by a set of characteristics that outline the requirements necessary for successful execution, while every agent is described through its Capability Profile. This dual representation allows us to rigorously compare the competencies of each agent against what is demanded by each task.

VBTA constructs a suitability matrix that quantitatively evaluates the compatibility between agents and tasks. For every agent-task pair, the matrix assigns a score reflecting how well the agent’s capabilities meet the task’s requirements. Notably, the highest-scoring agent is often the most qualified candidate for multiple tasks, necessitating a strategy to ensure that other agents are available to cover the remaining tasks.

To resolve potential conflicts in assignments, VBTA employs a robust voting and allocation mechanism. This system leverages six distinct voting methods along with three allocation strategies to balance conflicts, aggregate, and interpret the scores from the suitability matrix. Agent-task pair scores are calculated based on the suitability of the agents \textit{Capability Profile} when compared with the \textit{Task Description}, thus agents will have a higher score for tasks they are well suited to. As a result, VBTA carefully judges which agent should be allocated to each task, ensuring that assignments are aligned with both the agents' capabilities and the tasks' requirements.

In cases where voting alone cannot determine an agent-task pairing, and there is no other more qualified agent to allocate to, VBTA integrates an LLM into the decision-making process. This is a useful feature when a task's description and an agent's capability profile do not indicate clear compatibility. The LLM utilizes its semantic understanding to interpret subtle nuances in both the task descriptions and agent profiles, thereby refining the suitability assessment and ensuring a more informed assignment. The LLM prompt is automatically generated using the task description and agent capability profile, and the component(s) whose compatibility is in question becomes the focus of the prompt, (e.g. if the speed of the agent is in question, the prompt asks if task A that requires agent speed of X can be completed satisfactorily by agent B who has a speed of Y) creating a direct and easy to interpret prompt for the LLM.

After the assignment phase, VBTA simulates the agents' path planning to their designated tasks using CBS. Path planning is the task of finding an optimal sequence of states to satisfy an objective. These states can be locations on a map if we want to plan movement, game states if we want to plan a combat sequence, or dialogue states if we want to plan a narrative. This simulation not only facilitates efficient route planning for task execution but also demonstrates the framework's potential to handle more complex scenarios, such as dynamic game states, by optimizing the agents' trajectories toward their objectives. All types of path planning can be utilized by the agents during gameplay.

Overall, VBTA is designed to be extensible to any task that can be articulated through a clear set of characteristics. This adaptability makes it particularly well-suited for a range of applications in HRI and autonomous agents in role-playing games, where precise task definitions and efficient, reliable robot allocations are paramount.

\section{VBTA in Role-Playing Games}

Games provide a unique environment for the application of VBTA due to their clearly defined rules and structured objectives. In many games, agents are equipped with predefined capability profiles that detail what they can do, and tasks are accompanied by explicit descriptions of the requirements needed for successful completion. This structured setup allows VBTA to efficiently assign agents to tasks by comparing their capabilities with the demands of each task using a suitability matrix.

For instance, in a role-playing game, different agents might be responsible for tasks such as combat, exploration, negotiation, or puzzle solving. Each agent’s capability profile–reflecting strengths in areas like attack, defense, stealth, or persuasion–can be matched against task descriptions that specify the necessary skills for that role. VBTA leverages this information to score the suitability of each agent for every task, ensuring that the most appropriate agent is selected while also accounting for scenarios where the top agent might be best suited for multiple roles.

In addition to task assignment, VBTA incorporates CBS to optimize plans within the game environment. When applied to physical spaces, such as maps or level layouts, CBS calculates optimal routes that allow agents to reach their target locations efficiently while avoiding obstacles and conflicts with other agents. When VBTA applies CBS to abstract game states, the ``paths'' are not spatial routes but sequences of strategic actions–whether in combat, trade negotiations, stealth missions, persuasion challenges, or interactive dialogues. For example, during a dynamic battle scenario, VBTA may assign agents to specific roles based on their capabilities. Concurrently, CBS can generate a conceptual battle plan ``path'' that outlines a series of actions tailored to the evolving combat situation. As player actions or environmental conditions change, CBS will dynamically update these action plans to ensure that the agents remain on the most effective strategic trajectory, the goal being agents that make the player feel as if they are playing with the actual characters in the game.

Integrating VBTA with CBS not only ensures optimal agent-to-task assignments based on clearly defined criteria and semantic reasoning, but also allows it to adapt to changing game states in real time. This dual capability enhances overall game immersion by guaranteeing that agents act in contextually relevant and strategically sound ways, regardless of whether they are navigating a physical environment or a complex, evolving game state, resulting in intelligent, procedurally generated agent actions in every scenario the player runs into.

\subsection{Example Use Cases}
\subsubsection{Dungeons \& Dragons: Dungeon Master Assistant}

D\&D is a fantasy role-playing game renowned for its immersive storytelling and player autonomy. In a typical campaign, a DM orchestrates the entire game world, managing NPCs, enemies, environmental challenges, and even subtle narrative cues. This role demands not only creativity but also the ability to adapt quickly to the evolving actions of the players. The VBTA framework can assist the DM by automating and optimizing many of these decisions, ensuring that gameplay remains both dynamic and engaging.

In this context, VBTA leverages the pre-set capability profiles and task descriptions to populate the game world with responsive agents. For example, a \textit{Persuasion NPC Capability Profile} can be pre-configured for characters whose role is to interact with the players–be it to negotiate, deceive, or otherwise influence their decisions. Similarly, an \textit{Enemy NPC Capability Profile} is available to generate adversaries. Once the template capability profiles for a game are created, new ones can be generated from them with or without customization, and for a game like D\&D, all of this information is already available in the 5th edition game guide which contains the baseline data for pre-set capability profiles, reducing the overhead for new campaign generation. By introducing minor variations into these profiles during each playthrough, VBTA enables a procedurally generated experience that keeps the game fresh and unpredictable because of the complexity and diversity of agents available.

All in-game dialogue is generated by an LLM that is fed both the character’s goals and motivations as defined in its capability profile and task description via an automatically generated prompt. This process creates a narrative blueprint that guides each NPC’s interactions, ensuring that dialogue is contextually relevant and consistent with the character’s persona. Simultaneously, the agent’s actions—such as moving, attacking, defending, or negotiating—are determined by combining the capability profile and task description with CBS. CBS operates on the current game state, calculating optimal plans that respond intelligently to player behavior and evolving situations. Moreover, both capability profiles and task descriptions are dynamically updated as the campaign progresses to refine narrative and tactical decisions.

By automating these processes through a hybrid voting-based method integrated with semantic reasoning from an LLM, VBTA can shoulder a significant portion of the DM’s workload. This enables the DM to focus on higher-level storytelling and creative decision-making, ultimately crafting a richer and more immersive experience for players. With VBTA taking over the tedious work and contributing to the creative narrative, full campaigns—from overarching narrative arcs to spontaneous battle encounters and routine interactions like trading—can be executed with a high degree of personalization and strategic depth.

\subsubsection{Baldur's Gate 3: Narrative-Driven Game Agent}

Baldur's Gate 3 is a narrative-intensive game where immersion and dynamic storytelling are paramount. In such games, every NPC or narrative agent embodies a rich history, distinct emotions, and personal motivations that evolve in response to the player's actions. With a potential 1.4 million words of dialogue, 140 hours of cutscenes, and over 17,000 unique paths available per playthrough, achieving true replayability demands an extraordinary level of narrative depth and complexity. The lengths a game studio must go to for a game to have true replayability for years are immense, both before the release of a game and years of consistent paid content updates after release.

Instead, by using procedural generation, unique narratives and game levels can be created with each new playthrough. Recent approaches that rely solely on an LLM for procedural narrative generation often fall short in these expansive environments. A single LLM can become overwhelmed by the continuous context switches and branching storylines inherent in such a detailed game world. This is where VBTA offers a distinct advantage. By leveraging structured capability profiles—the same as those used in D\&D 5th edition game guide—and tailored task descriptions, VBTA can manage multiple narrative roles with precision.

For instance, narrative-driven agents that aim to motivate, deceive, or even antagonize the player can be assigned specific capability profiles that outline their character traits and behavioral tendencies. VBTA utilizes a suitability matrix and a voting mechanism to decide which narrative actions best align with the current game state, allowing the system to generate situation outlines that guide each agent's dialogue and decisions. In effect, rather than following a static, predetermined branching storyline, the narrative adapts dynamically. Each playthrough can yield a unique story as the agents’ responses, determined by a hybrid voting-based process employing an LLM, continuously update to reflect the evolving circumstances and player choices.

In summary, by integrating VBTA with an LLM and CBS for action planning, we can provide a robust framework that supports the immense narrative demands of games like D\&D and Baldur's Gate 3. This approach not only meets the technical requirements of complex, branching dialogues but also enhances player immersion by offering a uniquely tailored storytelling experience with every session. Where traditional branching narratives have to have many routes through the story pre-planned out, VBTA could dynamically build a narrative by creating a story capability profile whose task is to link all smaller narratives together, creating dynamic prompts for the LLM to fill in the gaps when needed, with new branches procedurally generated every time a player makes another choice.

\section{Conclusion}

VBTA is a flexible, generalizable framework. While procedural generation in role-playing games is one application, any well-defined multi-agent setting within HRI could be explored as a future use case. VBTA holds significant promise for transforming narrative-driven games by dynamically responding to both player emotions and the evolving game world. By leveraging a hybrid voting-based method to direct action and dynamically prompt an LLM for dialogue generation, VBTA can craft an almost limitless array of narrative paths, and interaction sequences, thus substantially increasing replayability and immersion. This approach not only reduces the development effort required to create complex titles like Baldur’s Gate 3 or personalized D\&D campaigns, but it also ensures that no matter how many times the player starts a new game, the subsequent dialogue and scenarios are uniquely generated because of the complexity and diversity of agents available to be spawned each playthrough. Current work is focused on evaluating VBTA's performance in simulation. Meanwhile, VBTA has the potential to create truly procedurally generated game agents, where every playthrough has stories and combat that can be both unpredictable and deeply engaging.

\bibliographystyle{ieeetr}
\bibliography{bibliography.bib}

\end{document}